\documentclass[10pt,twocolumn,letterpaper]{article}

\usepackage{cvpr}              % To produce the CAMERA-READY version
\usepackage[dvipsnames]{xcolor}

\usepackage{graphicx}
\usepackage{amsmath}
\usepackage{amssymb}
\usepackage{booktabs}
\usepackage{xcolor}
\usepackage{tikz}
\definecolor{gold}{HTML}{BD820B}%{EEAD0E}
\definecolor{silver}{HTML}{909090}%{C0C0C0}
\definecolor{bronze}{HTML}{9A5F26}%{CD7F32}
\usepackage{color, colortbl}
\definecolor{Gray}{gray}{0.95}

\newcommand*\circledd[1]{\tikz[baseline=(char.base)]{
            \node[shape=circle,draw,inner sep=0.15pt] (char) {#1};}}      

\newcommand{\first}[1]{%
    {#1\raisebox{0.8pt}{\footnotesize \color{gold} \circledd{1}}}%
}
\newcommand{\second}[1]{%
    {#1\raisebox{0.8pt}{\footnotesize \color{silver} \circledd{2}}}%
}
\newcommand{\third}[1]{%
    {#1\raisebox{0.8pt}{\footnotesize \color{bronze} \circledd{3}}}%
}

\makeatletter
\newcommand\footnoteref[1]{\protected@xdef\@thefnmark{\ref{#1}}\@footnotemark}
\makeatother

\definecolor{cvprblue}{rgb}{0.21,0.49,0.74}
\usepackage[pagebackref,breaklinks,colorlinks,citecolor=cvprblue]{hyperref}

\graphicspath{{./figures/}}
\def\paperID{*****} % *** Enter the Paper ID here
\def\confName{CVPR}
\def\confYear{2024}

\title{DAVE -- A Detect-and-Verify Paradigm for Low-Shot Counting}

\author{Jer Pelhan, Alan Lukežič, Vitjan Zavrtanik, Matej Kristan \\
\small {Faculty of Computer and Information Science, University of Ljubljana, Slovenia}\\
{\tt\small jer.pelhan@fri.uni-lj.si}}
\begin{document}
\maketitle

\begin{abstract}
Low-shot counters estimate the number of objects corresponding to a selected category, based on only few or no exemplars annotated in the image. 
The current state-of-the-art estimates the total counts as the sum over the object location density map, but does not provide individual object locations and sizes, which are crucial for many applications. This is addressed by detection-based counters, which, however
fall behind in the total count accuracy.
Furthermore, both approaches tend to overestimate the counts in the presence of other object classes due to many false positives.
We propose DAVE, a low-shot counter based on a detect-and-verify paradigm, that avoids the aforementioned issues by first generating a high-recall detection set and then verifying the detections to identify and remove the outliers. 
This jointly increases the recall and precision, leading to accurate counts. 
DAVE outperforms the top density-based counters by $\sim$20\% in the total count MAE, it outperforms the most recent detection-based counter by $\sim$20\% in detection quality and sets a new state-of-the-art in zero-shot as well as text-prompt-based counting.
The code and models are available on \href{https://github.com/jerpelhan/DAVE}{GitHub}.
\end{abstract}

\vspace{-0.4em}
\section{Introduction}\label{sec:intro}

%\cmnt{[MK] Nenavadno je uporabljati Fig., Sec. Tab. -- ker imamo referenciranje avtomatko, uporabimo ukaze, da bo pisalo Figure, Section, Table.}
\vspace{-0.4em}

Low-shot counting considers estimating the number of target objects in an image, based only on a few annotated exemplars (few-shot) or even without providing the exemplars (zero-shot). 
% In the latter scenario, the algorithm has to report the number of the majority-class objects. 
Owing to the emergence of focused benchmarks~\cite{famnet,counting-detr}, there has been a surge in low-shot counting research recently. The current state-of-the-art low-shot counters are all density-based~\cite{Shi_2022_CVPR, you2023few, famnet, djukic_loca}. This means that they estimate the total count by summing over an estimated object presence density map. Only recently, few-shot detection-based methods emerged~\cite{counting-detr} that estimate the counts as the number of detected objects.

%estimate the counts as the number of detected objects.

Density-based methods substantially outperform the detection-based counters in total count estimation, but they do not provide detailed outputs such as object locations and sizes. The latter are however important in many downstream tasks such as bio-medical analysis~\cite{zavrtanik2020segmentation,xie2018microscopy}, where explainability is crucial for human expert verification as well as for subsequent analyses. There is thus a large applicability gap between the density-based and detection-based low-shot counters.

 \begin{figure}[t]

  \centering
  \includegraphics[width=0.47\textwidth]{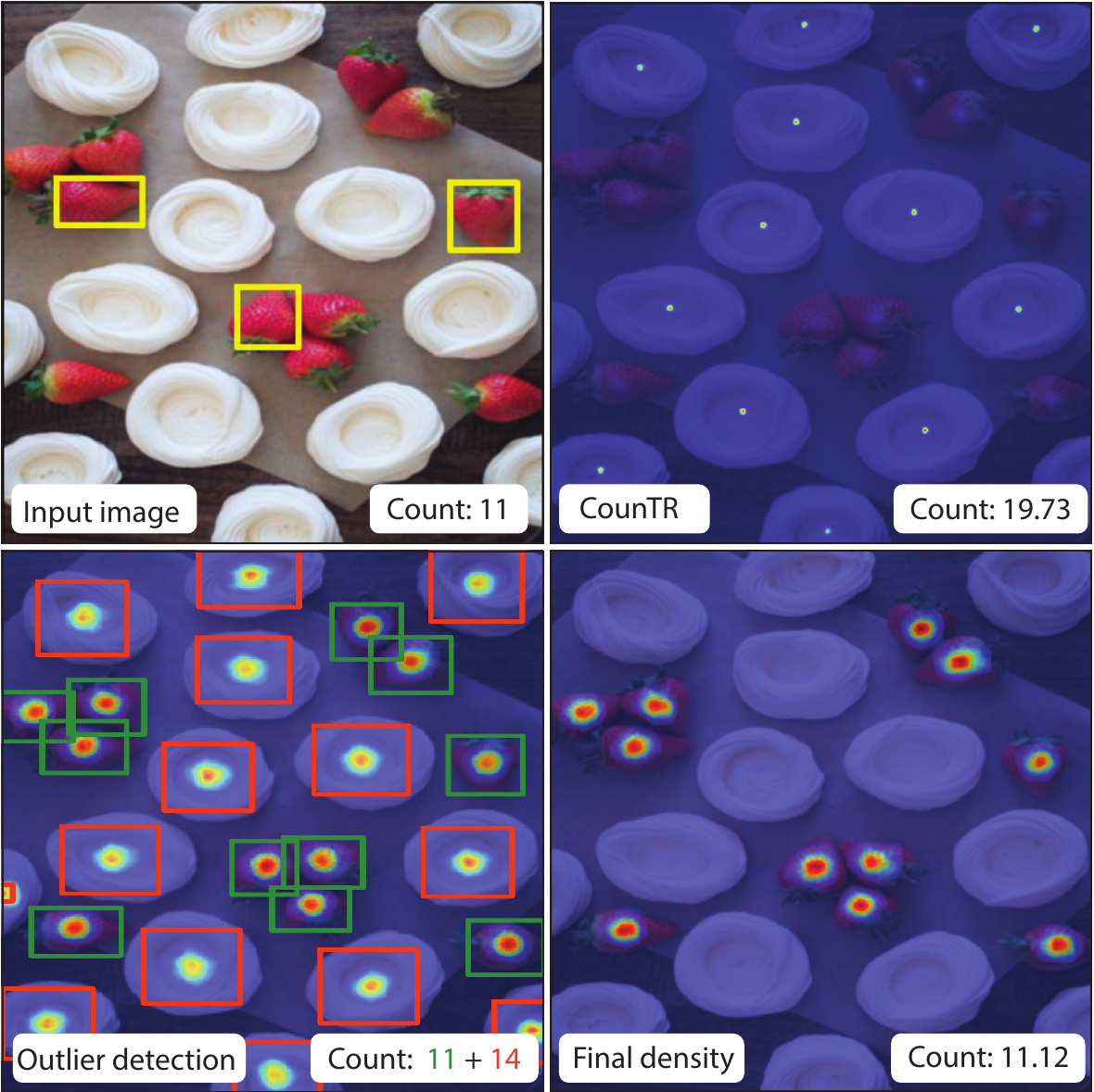} 
  \caption{
  Despite considering exemplars (yellow boxes), the state-of-the-art (e.g., CounTR~\cite{Liu_2022_BMVC}) is prone to false activations on incorrect objects, leading to corrupted counts. DAVE avoids this issue by \textit{detecting} all candidates (red and green boxes), \textit{verifying} them, removing the outliers (red boxes), and correcting the final density map, thus jointly improving detection and count estimation.
  }
\label{fig:motivation}
\end{figure}
% Existing class-agnostic counters are prone to failure due to false activations on outliers. DAVE avoids this by generating a high-recall detection set and then removing the outliers in the verification step, thus producing accurate detections and as well as improving the density-based total count estimation.

Furthermore, both density-based and detection-based counters are prone to failure in scenes with several object types (\Cref{fig:motivation}). The reason lies in the specificity-generalization tradeoff. Obtaining a high recall requires generalizing over the potentially diverse appearances of the selected object type instances in the image. 
However, this also leads to false activations on objects of other categories (false positives), leading to a reduced precision and count overestimation. 
A possible solution is to train on multiple-class images~\cite{counting-detr}, however, this typically leads to a reduced recall and underestimated counts.

%However, this also leads to false activations on objects of other categories, leading to a reduced precision. The estimated counts are thus corrupted by false positives. 
%A possible solution is to train on multiple-class images~\cite{counting-detr}, however, this typically leads to a reduced recall and underestimated counts.

We address the aforementioned issues by proposing a low-shot counter DAVE, which combines the benefits of density-based and detection-based formulations, and introduces a novel \underline{d}etect-\underline{a}nd-\underline{ve}rify paradigm.
DAVE tackles the specificity-generalization issues of the existing counters by applying a two-stage pipeline~(\Cref{fig:motivation}). In the first, \textit{detection} stage, DAVE leverages density-based estimation to obtain a high-recall set of candidate detections, which however may contain false positives. This is addressed by the second, \textit{verification} stage, where outliers are identified and rejected by analyzing the candidate appearances, thus increasing the detection precision. 
Regions corresponding to the outliers are then removed from the location density map estimated in the first stage, thus improving the density-based total count estimates as well.
In addition, we extend DAVE to text-prompt-based and to a zero-shot scenario, which makes DAVE the first zero-shot as well as text-prompt detection-capable counter.
%thus covering the entire range of low-shot counting setups, which makes DAVE the first low-shot counter capable also of zero-shot counting and detection.

{
The primary contribution of the paper is the detect-and-verify paradigm for low-shot counting that simultaneously achieves high recall and precision. 
The proposed architecture is the first to extend to all low-shot counting scenarios. 
DAVE uniquely merges the benefits of both density and detection-based counting and is the first zero-shot-capable counter with detection output. 
}
DAVE outperforms all state-of-the-art density-based counters on the challenging benchmark~\cite{famnet}, including the longstanding winner~\cite{djukic_loca}, achieving a 
relative 20\% MAE and 43\% RMSE total-count error reductions. 
It also outperforms all state-of-the-art detection-based counters on the recent benchmark FSCD147~\cite{counting-detr} by $\sim$20\% in detection metrics, as well as in the total count estimation by 38\% MAE.
Furthermore, it sets a new state-of-the-art in text-prompt-based counting.
The zero-shot DAVE variant outperforms all zero-shot density-based counters and delivers detection accuracy on-par with the most recent \textit{few-shot} counters. 
DAVE thus simultaneously outperforms both density-based and detection-based counters in a range of counting setups.

%%%%%%%%%%%%%%%%%%%%%%%%%%%%%%%%%%%%%%%%%%%%%%%%%%%%%%%%%%%%%%%%%%%%%%%%%%%%%%%%
\section{Related Work}
\vspace{-0.4em}
Object counting emerged as detection-based counting of objects belonging to specific classes, such as vehicles~\cite{vehicle-counting}, cells~\cite{cell-counting-detection}, people~\cite{crowd-counting}, and polyps~\cite{zavrtanik2020segmentation}.
To address poor performance in densely populated regions, density-based methods~\cite{shu2022crowd,cheng2022rethinking,wan2020modeling,wang2020distribution, Cholakkal_2019_CVPR} emerged as an alternative. 

All these methods rely on the availability of large datasets to train category-specific models, which, however are not available in many applications.

Class-agnostic approaches addressed this issue by test-time adaptation to various object categories with minimal supervision. 
Early representatives~\cite{first-fsc} and~\cite{yang2021class} 
proposed predicting the density map by applying a siamese matching network to compare image and exemplar features.
Recently, the FSC147 dataset~\cite{famnet} was proposed to encourage the development of few-shot counting methods. 
Famnet~\cite{famnet} proposed a test-time adaptation of the backbone to improve density map estimation.
BMNet+~\cite{Shi_2022_CVPR} improved localization by jointly learning representation and a non-linear similarity metric. A self-attention mechanism was applied to reduce the intra-class appearance variability. 
SAFECount~\cite{you2023few} introduced a feature enhancement module, improving generalization capabilities.
CounTR~\cite{Liu_2022_BMVC} used a vision transformer~\cite{vit} for image feature extraction and a convolutional encoder to extract exemplar features. An interaction module based on cross-attention was proposed to fuse both, image and exemplar features. 
LOCA~\cite{djukic_loca} proposed an object prototype extraction module, which combined exemplar appearance and shape with an iterative adaptation. 

All few-shot counting methods require few annotated exemplars to specify the object class. 
With the recent development of large language models (e.g.~\cite{clip}) text-prompt-based counting methods emerged. 
Instead of specifying exemplars by bounding box annotations, these methods use text descriptions of the target object class. % to specify the object class. 
ZeroCLIP~\cite{xu2023zero} proposed text-based construction of prototypes, which are used to select relevant image patches acting as exemplars for counting.
CLIPCount~\cite{jiang2023clip} leveraged CLIP~\cite{clip} for image-text alignment and introduced patch-text contrastive loss for learning the visual representations used for density prediction.
Several works~\cite{ranjan2022exemplar,hobley2022learning} address the extreme case in which no exemplars are provided and the task is to count the majority class objects (i.e., zero-shot counting).  

With minimal architectural changes, the recent few-shot methods~\cite{Liu_2022_BMVC,djukic_loca} also demonstrated a remarkable zero-shot counting performance. A common drawback of density-based counters is that they do not provide object locations.

To address the aforementioned limitation of density-based counters, the first few shot counting and detection method~\cite{counting-detr} has been recently proposed by extending a transformer-based object detector~\cite{carion2020end} with an ability to detect objects specified by exemplars. However, the detection-based counter falls far behind in total count estimation compared with the best density-based counters.

%%%%%%%%%%%%%%%%%%%%%%%%%%%%%%%%%%%%%%%%%%%%%%%%%%%%%%%%%%%%%%%%%%%%%%%%%%%%%%%%
\section{Counting by detection and verification}  \label{sec:dave}
%%%%%%%%%%%%%%%%%%%%%%%%%%%%%%%%%%%%%%%%%%%%%%%%%%%%%%%%%%%%%%%%%%%%%%%%%%%%%%%%
\vspace{-0.4em}
\begin{figure*}[t]
  \centering
  \includegraphics[width=\textwidth]{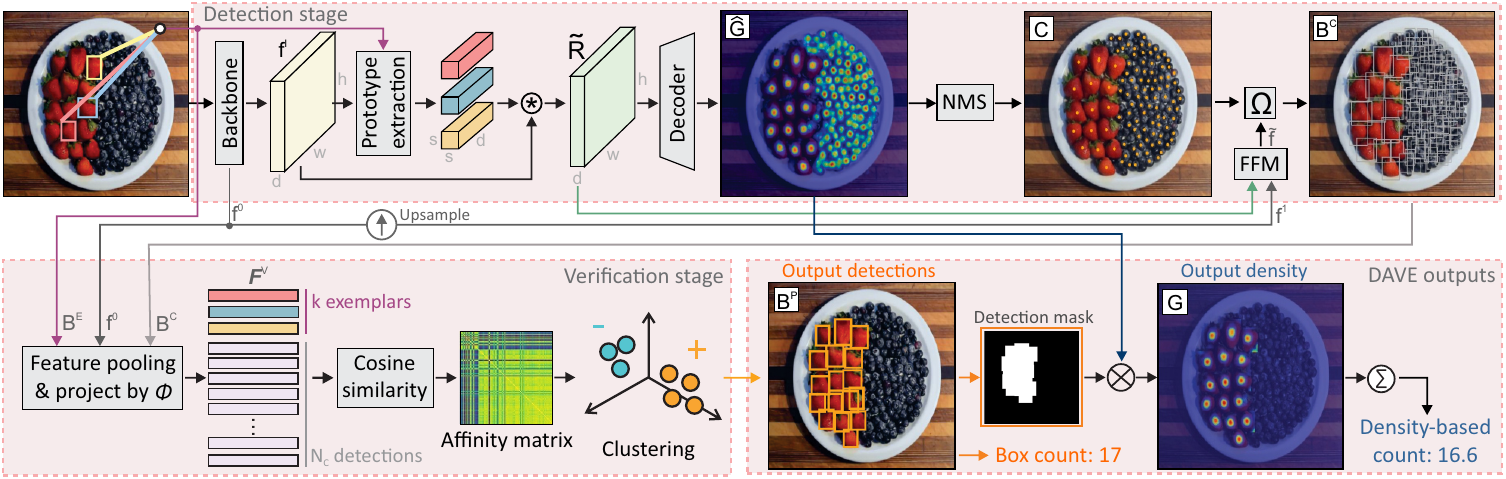}

  \caption{The proposed DAVE architecture consists of two stages, (i) {\it detection} and (ii) {\it verification}, and outputs detected objects as well as an improved location density map. 
  NMS denotes non-maxima suppression, FFM is a feature fusion module, $\Omega$ is a bounding box regression head and $\phi$ is the verification feature extraction network.
  }  \label{fig:dave_detailed}
\end{figure*}

Formally, given an input image $I \in \mathbb{R}^{H_0 \times W_0 \times 3}$ and a set of $k$ exemplar bounding boxes $\boldsymbol{B}^{\mathrm{E}} = \{ b_i \}_{i=1:k}$ denoting object exemplars, a low-shot detection counter is required to report bounding boxes $\boldsymbol{B}^{P} = \{ b_i \}_{i=1:N_P}$ of all detected objects of the same category and their estimated count. 

In the following we present the new \underline{d}etect-\underline{a}nd-\underline{ve}rify few-shot counting and detection method (DAVE), which consists of two stages.
In the first, {\it detection}, stage (Section~\ref{sec:detect}), 
candidate regions are estimated by pursuing a high recall, potentially including false positive detections, i.e., objects belonging to an incorrect category.
In the second, {\it verification}, stage (Section~\ref{sec:verify}) 
the candidate regions are analyzed to identify and reject the outliers, thus increasing the detection precision.
The outliers are used to update the density map, thus also improving density-based count estimation, which may compensate for missed objects and can differ from the number of detections $N_P$.
A detailed DAVE architecture is shown in Figure~\ref{fig:dave_detailed}.

%-----------------------------------------------------------------------------

\subsection{Detection stage}  \label{sec:detect}
\vspace{-0.4em}
The aim of this stage is to predict candidate bounding boxes $\boldsymbol{B}^{C} = \{ b_i \}_{i=1:N_c}$ with a high recall.
Detection is thus split into first estimating the object centers $\boldsymbol{C} = \{ (x_c^{i}, y_c^{i}) \}_{i=1:N_c}$, and then predicting the corresponding bounding box parameters.
We re-purpose the architecture of the recent low-shot counter LOCA~\cite{djukic_loca} for estimating the object location density map $\mathbf{\tilde{G}}$, from which we obtain the center locations $\boldsymbol{C}$ by non-maxima suppression.\footnote{{The minimal distance between two peaks in NMS is set to 1 to maximize the detection rate.}}
Briefly, the location density estimation architecture is the following: the input image is first encoded by ResNet-50~\cite{he2016deep}, followed by a transformer, generating the representation $\mathbf{f}^{I} \in \mathbb{R}^{h \times w \times d}$.
Exemplar prototypes are constructed using the OPE module~\cite{djukic_loca} and correlated with $\mathbf{f}^{I}$ resulting in a similarity tensor $\mathbf{\tilde{R}} \in \mathbb{R}^{h \times w \times d}$.
A decoder is then applied on $\mathbf{\tilde{R}}$ to obtain the final 2D location density map $\mathbf{\hat{G}} \in \mathbb{R}^{H_0 \times W_0}$. We refer the reader to~\cite{djukic_loca} for more details.

Next, features are constructed for regressing the bounding box parameters for each detected center. 
The feature construction pipeline is designed to ensure that final features reflect the objectness information specific to the class selected by the exemplars.
Features from the second, third, and fourth blocks of the backbone are resized to $64 \times 64$ pixels, concatenated along the channel dimension and reduced to $d$ channels using a $3 \times 3$ convolution, i.e. $\mathbf{f}^0 \in \mathbb{R}^{h \times w \times d}$. The features are then upsampled to match the input image size ($\mathbf{f}^1 \in \mathbb{R}^{H_0 \times W_0 \times d}$). 
Next, %the selected category objectness information
the selected object category shape information 
is injected by fusing $\mathbf{f}^1$ and the upscaled similarity tensor $\mathbf{\tilde{R}}$ using the feature fusion module (FFM)~\cite{yu2018bisenet}, i.e., $\tilde{\mathbf{f}} = \mathrm{FFM}(\mathbf{f}^1, \tilde{\mathbf{R}})$.

The constructed features $\tilde{\mathbf{f}}$ are then fed into a bounding box regression head $\Omega(\cdot)$ akin to~\cite{yu2016unitbox,huang2015denseboxtlrb}, which predicts for each location a distance to the left, right, top and bottom bounding box edge of the underlying object. The network 
$\Omega(\cdot)$ consists of two $3 \times 3$ convolutional layers with $d$ and $4$ channels with GroupNorm~\cite{groupnorm} and ReLU operations in between, and predicts a dense bounding box map $\mathbf{v} \in \mathbb{R}^{H_0 \times W_0 \times 4}$.
The object candidate bounding boxes $\boldsymbol{B}^{C}$ are thus obtained by reading out the corresponding values from $\mathbf{v}$ at locations $\boldsymbol{C}$.

%-----------------------------------------------------------------------------
\subsection{Verification stage}  \label{sec:verify}
\vspace{-0.4em}
In practice, the candidate detections $\boldsymbol{B}^{C}$ retain a high recall, but are also contaminated by false positives. 
The goal of the verification stage is thus to increase the precision by analysing the appearance of the detections and rejecting the outliers. 
First, a verification feature vector $\mathbf{f}^{\mathrm{v}}_i$ is extracted for each detected bounding box $b_i$ as follows.
The backbone features $\mathbf{f}^{0}$ are pooled into a feature tensor $\mathbf{f}_i \in \mathbb{R}^{s \times s \times d}$ and transformed by a shallow network $\phi(\cdot)$ consisting of two $1 \times 1$ convolutions with $d$ channels and a BatchNorm and ReLu activation in between. 
The verification features are also extracted for the annotated exemplars, leading to $N_C + k$ features in total, i.e., $\boldsymbol{F}^{\mathrm{V}} = \{ \mathbf{f}^{\mathrm{v}}_i \}_{i=1:(N_C + k)}$.

The verification features are then clustered by unsupervised clustering. Specifically, spectral clustering~\cite{spectral_clustering} is applied to an affinity matrix computed from cosine similarities between pairs of features in $\boldsymbol{F}^{\mathrm{V}}$, yielding several clusters. 
Object candidate detections belonging to clusters with at least one exemplar are kept, while the other are labelled as outliers and removed, yielding the final set of $N_P$ object detections $\boldsymbol{B}^{P} = \{ b_i \}_{i=1:N_P}$. 
Finally, the density map $\mathbf{\hat{G}}$ from the detection stage is updated by setting all values outside of the detected bounding boxes to zero, yielding $\mathbf{G}$, from which the improved density-based count is estimated (\Cref{fig:dave_detailed}).

%-----------------------------------------------------------------------------
\subsection{Zero-shot and prompt-based adaptation}  \label{sec:zero-adapt}

{\textbf{Zero shot counting.}} DAVE is easily adapted to a zero-shot setup in which exemplars are not provided and the task is to count and detect the majority-class objects. First, the location density prediction part is replaced by its zero-shot variant~\cite{djukic_loca} to account for the absence of exemplars. The detection stage and most of the verification stage remain unchanged. The only change in the verification stage is the cluster selection method: all clusters whose size is at least 45\%\footnote{\label{note1}{Extensive analysis shows robustness to this hyperparameter.}} of the largest cluster are kept as positive detections and the rest are identified as outliers. This is to account for the possibility that clusters may break up due to the absence of exemplars specifying the level of appearance similarity.

{\textbf{Prompt-based counting.}}
%The zero-shot DAVE is extended to the prompt-based counting setup, in which the target object class is specified by a text prompt. The only modification is the cluster selection protocol in the verification stage. 
%For each cluster, a CLIP~\cite{radford2021learning} embedding is extracted from the image with all values outside of the corresponding bounding boxes set to zero and compared with the CLIP~\cite{radford2021learning} text prompt embedding. 
%All clusters with less than 85\% of the highest prompt-to-cluster similarity are identified as outliers. 
Zero-shot DAVE is extended to the prompt-based counting setup, in which the target object class is specified by a text prompt. The only modification is the cluster selection protocol in the verification stage. 
The text prompt embedding is extracted by CLIP and compared to the CLIP embedding of each identified cluster. The latter is obtained by masking the image regions outside the bounding boxes corresponding to the cluster and computing the CLIP embedding. Cosine distances between the text embedding and individual cluster embeddings are computed, and clusters with less than 85\%\footref{note1} of the highest prompt-to-cluster similarity are identified as outliers.

\subsection{Training}
\label{sec:training}
\vspace{-0.4em}
Few-shot counting datasets typically contain centers of all objects annotated and the bounding boxes available for only $k=3$ exemplars. We formulate the training to adhere to these restrictions. The object centers can be used to train location density prediction network. Since DAVE employs LOCA~\cite{djukic_loca} for the initial density prediction, we use the publicly available pretrained version of LOCA, and train only the free parameters of the detection and verification stages in two phases. 

In the first phase, the detection stage (i.e., the FFM and $\Omega(\cdot)$) is trained by a bounding box loss evaluated on the available ground truth exemplar bounding boxes, i.e.,  
 $\mathcal{L}_{box} = \sum\nolimits_{i=1}^{k=3} 1 - \mathrm{GIoU}(\mathbf{v}(x^{c}, y^{c}), b_i^{\mathrm{GT}})$,
% \begin{equation}
% \mathcal{L}_{box} = \sum_{i=1}^{k=3} 1 - \mathrm{GIoU}(\mathbf{v}(x^{c}, y^{c}), b_i^{\mathrm{GT}}),
% \end{equation}
where $(x^{(i)}_c, y^{(i)}_c)$ are locations in the central regions of the ground truth bounding boxes $b_i^{\mathrm{GT}}$ and $ \mathrm{GIoU}(\cdot)$ is the generalized intersection over union~\cite{giou_cvpr2018}

In the second phase, the verification feature extraction network $\phi(\cdot)$ is trained. Training examples are generated by stitching together a pair of images with annotated exemplar objects of different classes. The stitched image thus contains $2 \times 3=6$ bounding boxes, yielding two sets of features extracted by $\phi(\cdot)$, corresponding to the two sets of exemplars: $\{ \mathbf{z}_j^1 \}_{j=1:3}$ and $\{ \mathbf{z}_j^2 \}_{j=1:3}$. The verification network $\phi(\cdot)$ is then trained by a contrastive loss~\cite{contrastive_loss}:
\[
    \mathcal{L}_{cos} = 
\begin{cases}
    1 - c(z_{j_1}^{i_1}, z_{j_2}^{i_2}),& i_1 = i_2 \\
    \mathrm{max}(0, c(z_{j_1}^{i_1}, z_{j_2}^{i_2}) - \lambda),              & \text{else},
\end{cases}
\]
where $c(z_{j_1}^{i_1}, z_{j_2}^{i_2})$ is the cosine similarity between a pair of features, and $\lambda$ is the margin.

%%%%%%%%%%%%%%%%%%%%%%%%%%%%%%%%%%%%%%%%%%%%%%%%%%%%%%%%%%%%%%%%%%%%%%%%%%%%%%%%
\section{Experiments}
%%%%%%%%%%%%%%%%%%%%%%%%%%%%%%%%%%%%%%%%%%%%%%%%%%%%%%%%%%%%%%%%%%%%%%%%%%%%%%%%

\subsection{Implementation details}
\label{sec:implementation_details}

\textbf{Preprocessing.} 
Following~\cite{Liu_2022_BMVC}, the input image is resized such that the mean of the exemplars width and height is between 50 and 10 pixels. 
In the zero-shot setup, the method is bootstrapped, with applying the first pass to estimate the object sizes and then applying the second pass with the resizing as in the few-shot case.

\textbf{Training.}  
%The training proceeds in two stages.
In the first training stage, the feature fusion module FFM and the box regression head $\Omega(\cdot)$ are trained for 50 epochs by AdamW~\cite{loshchilov2017decoupled} with learning rate $10^{-4}$, weight decay $10^{-4}$ and batch size 8. 
The size of input images is kept fixed ($H_0 = W_0 = 512$) by zero-padding. 
In the second stage, the verification feature extraction network $\phi(\cdot)$ is trained for 50 epochs by AdamW~\cite{loshchilov2017decoupled} with the learning rate  $10^{-5}$, the weight decay $10^{-4}$, and the batch size 64.

\subsection{Density-based counting performance} \label{sec:counting_perf}
\vspace{-0.4em}
%We first compare 
DAVE is compared with the density-based state-of-the-art counters. 
For consistent comparison, density-based count estimation is considered in DAVE as well, i.e., the count is estimated by summation of the output location density map ${\mathbf{G}}$ (Section~\ref{sec:verify}).
The methods are evaluated on the challenging FSC147~\cite{famnet}, which contains 6135 images of 147 object classes, split into 3659 training, 1286 validation and 1190 test images. The object classes are disjoint across the splits to reflect realistic applications where the target object category is unseen during training. In each image, three exemplars are annotated with bounding boxes and all target objects by point annotations. The standard evaluation protocol~\cite{famnet,Shi_2022_CVPR,you2023few}  with Mean Absolute Error (MAE) and Root Mean Squared Error (RMSE) is followed.

\textbf{Few-shot counting.} In few-shot counting, all three exemplars are considered as the input. DAVE is compared with the most recent state-of-the-art density-based counters: LOCA~\cite{djukic_loca}, CounTR~\cite{Liu_2022_BMVC}, SAFECount~\cite{you2023few}, BMNet+~\cite{Shi_2022_CVPR},  VCN~\cite{Ranjan_2022_CVPR}, CFOCNet~\cite{yang2021class}, MAML~\cite{finn2017model}, FamNet~\cite{famnet}, and CFOCNet~\cite{yang2021class}. Results are summarized in \Cref{tab:fsc147-results}. 

DAVE outperforms all few-shot density-based counters by a large margin. It outperforms the long-standing winner LOCA~\cite{djukic_loca} 
by 13\% and 20\% in MAE on validation and test sets, respectively. It achieves a relative improvement of 14\% and a remarkable 43\% RMSE on the validation and test sets, respectively, setting a solid new state-of-the-art. 

\begin{table}[ht]
\centering
\caption{
Few-shot density-based counting on the FSC147~\cite{famnet}.
}
\label{tab:fsc147-results}
\scalebox{0.93}{
\begin{tabular}{lllll}
\toprule
 & \multicolumn{2}{c}{Validation set} & \multicolumn{2}{c}{Test set} \\
\cmidrule(lr){2-3} \cmidrule(lr){4-5}
Method & MAE & RMSE & MAE  & RMSE  \\
\midrule
GMN~\cite{lu2019class} & 29.66 & 89.81 & 26.52 & 124.57 \\
MAML~\cite{finn2017model} & 25.54 & 79.44 & 24.90 & 112.68 \\
FamNet~\cite{famnet} & 23.75 & 69.07 & 22.08 & 99.54 \\
CFOCNet~\cite{yang2021class} & 21.19 & 61.41 & 22.10 & 112.71 \\

BMNet+~\cite{Shi_2022_CVPR} & 15.74 & 58.53 & 14.62 & 91.83 \\

VCN~\cite{Ranjan_2022_CVPR} &19.38& 60.15& 18.17& 95.60 \\
SAFECount~\cite{you2023few} & 15.28  & 47.20\third{}  & 14.32  & 85.54\third{}  \\
CounTR~\cite{Liu_2022_BMVC} & 13.13\third{}  & 49.83 & 11.95\third{}  & 91.23  \\

LOCA~\cite{djukic_loca} & 10.24\second{}  & 32.56\second{}  & 10.79\second{}  & 56.97\second{}  \\
DAVE &  8.91\first{}   & 28.08\first{} & 8.66\first{} & 32.36\first{} \\
% \midrule
% \midrule
% C-DETR~\cite{counting-detr}$\ast$& 20.38& 82.45&16.79& 123.56 \\
% DAVE$^{\text{box}}$ $\ast$ & 9.71  & 32.59 & 9.34  & 50.27  \\
\bottomrule
\end{tabular}
}
\end{table}

\begin{figure*}[ht]
  \centering
  \includegraphics[width=\textwidth]{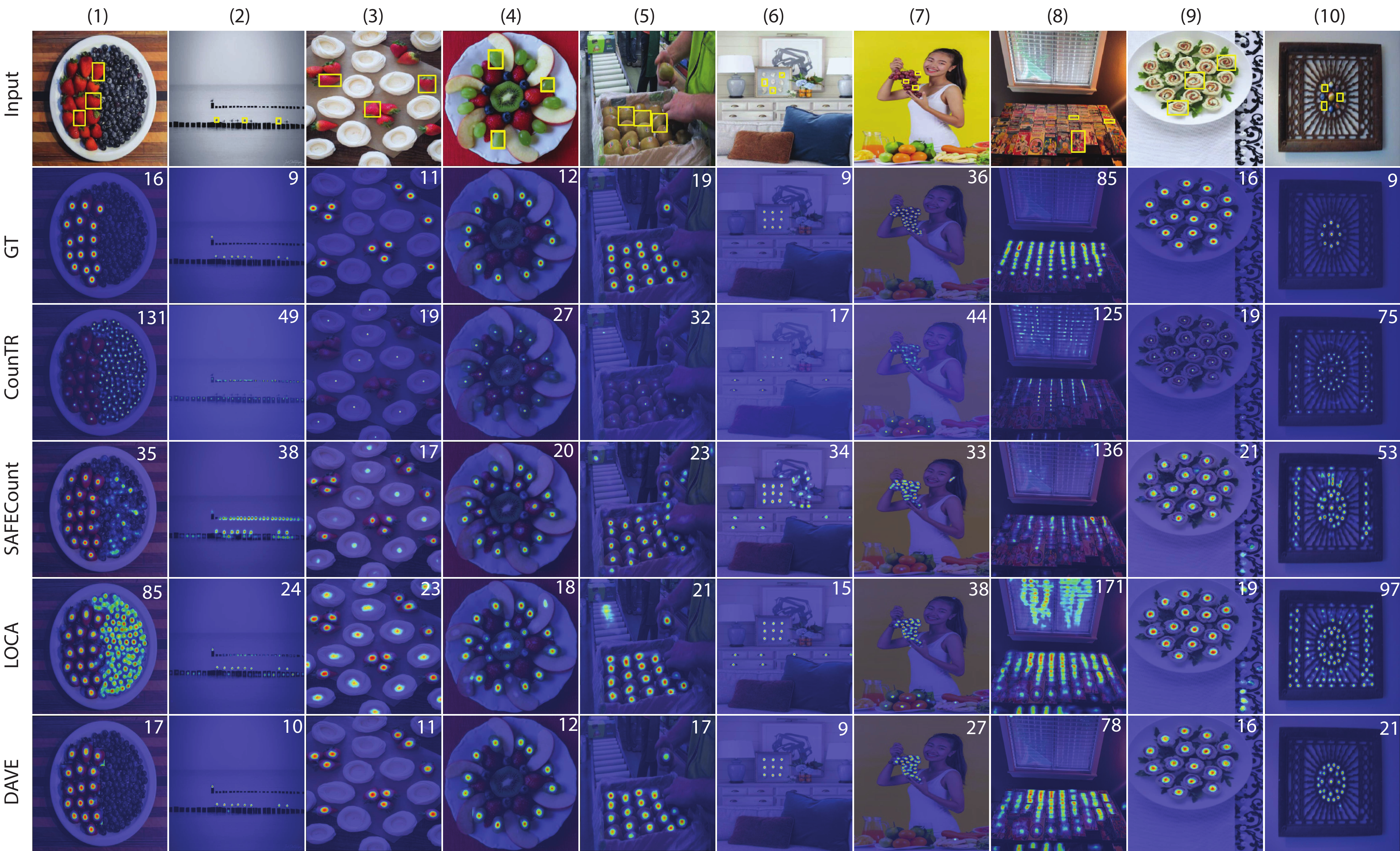} 
  \caption{Qualitative comparison of DAVE with LOCA~\cite{djukic_loca}, SAFECount~\cite{you2023few} and CounTR~\cite{Liu_2022_BMVC}. The first two columns show the input images and the ground truth (GT), while the predicted densities are shown in the rest.
  }
  \label{fig:qualitative_cnt}
\end{figure*}

\vspace{-1em}
To verify the source of performance improvements, we visualize DAVE density predictions and compare them with the recent state-of-the-art methods (\Cref{fig:qualitative_cnt}). 
We observe that other methods often count objects of an incorrect category (columns 1, 2, 3, 4, 5, 6, 7) or structures in the background texture (columns 8, 9, 10).
This indicates that related methods over-generalize localization features, which increases the recall at the cost of reduced precision.
DAVE, however, retains the high recall, while successfully identifying the outliers and suppressing the corresponding activations in the density map, thus improving precision. 
This indicates the strong benefits of the proposed detect-and-verify paradigm for density-based counting.

\textbf{One-shot counting.} In the one-shot counting setup, a single exemplar is considered. Comparison with the recent state-of-the-art methods GMN~\cite{lu2019class}, CFOCNet~\cite{yang2021class}, FamNet~\cite{famnet}, BMNet+~\cite{Shi_2022_CVPR}, CounTR~\cite{Liu_2022_BMVC}, and LOCA~\cite{djukic_loca} is reported in \Cref{tab:fsc147-1shot-results}. 
DAVE excels in one-shot counting, surpassing the previous best-performing methods by 5\% and 10\% MAE, and 9\% and 12\% RMSE on the validation and test set, respectively. Results indicate that the detect-and-verify paradigm helps to fully utilize the meaningful information from the only available exemplar, leading to performance improvements.
%This further confirms the benefits of the detect and verify paradigm even in one-shot counting scenario.
\begin{table}[ht]
\centering
\caption{
One-shot density-based counting on the FSC147~\cite{famnet}.
}
\label{tab:fsc147-1shot-results}
\resizebox{1 \linewidth}{!}{
\begin{tabular}{lllll}
\toprule
 & \multicolumn{2}{c}{Validation set} & \multicolumn{2}{c}{Test set} \\
\cmidrule(lr){2-3} \cmidrule(lr){4-5}
Method & MAE & RMSE & MAE  & RMSE  \\
\midrule
GMN~\cite{lu2019class} & 29.66 & 89.81 & 26.52 & 124.57 \\
CFOCNet~\cite{yang2021class} & 27.82 & 71.99 & 28.60 & 123.96 \\
FamNet~\cite{famnet} & 26.55 & 77.01 & 26.76 & 110.95 \\
BMNet+~\cite{Shi_2022_CVPR} & 17.89 & 61.12 & 16.89 & 96.65 \\
CounTR~\cite{Liu_2022_BMVC} & 13.15\third{} & 49.72 \third{}& 12.06\second{} & 90.01\third{} \\
LOCA~\cite{djukic_loca} & 11.36\second{} & 38.04\second{} & 12.53\third{} & 75.32\second{} \\
DAVE  & 10.79\first{}& 34.55\first{} & 11.29\first{} & 66.36\first{}\\
\bottomrule
\end{tabular}
}
\end{table}

%\subsubsection{Zero-shot counting} 

\textbf{Prompt-based counting.} 
The prompt-based modification of DAVE from \Cref{sec:zero-adapt} (denoted here as DAVE$_{\text{prm}}$)
is compared with the recent state-of-the-art prompt-based counters ZeroClip~\cite{xu2023zero}, CounTX~\cite{CounTX} and CLIP-Count~\cite{jiang2023clip}. Results in~\Cref{tab:prompt} show that DAVE$_{\text{prm}}$ outperforms the best counter (CounTX~\cite{CounTX})
by 12\% and 5\% MAE and 17\% and 3\% RMSE on validation and test sets, respectively. DAVE thus sets a solid new state-of-the-art in this setup.
%, without relying on the additional information provided by text.

\begin{table}[ht]
    \centering
    \caption{  
    Prompt-based counting on the FSC147~\cite{famnet}. 
    }
    \label{tab:prompt}
    \scalebox{0.9}{
    \begin{tabular}{lllll}
         \toprule
         & \multicolumn{2}{c}{Validation Set} & \multicolumn{2}{c}{Test Set} \\
        \cmidrule(lr){2-3} \cmidrule(lr){4-5}
       Method  & MAE  & RMSE & MAE  & RMSE \\
        \midrule

       ZeroClip~\cite{xu2023zero} &26.93& 88.63 &22.09 &115.17 \\
       CLIP-Count~\cite{jiang2023clip} &18.79\third{}&	61.18\second{}&17.78\third{}{}	&106.62\second{} \\
       CounTX~\cite{CounTX}   & 17.70\second{} & 63.61\third{} & 15.73\second{} & 106.88\third{} \\
        DAVE$_{\text{prm}}$ &15.48\first{}&52.57\first{}&14.90\first{}&103.42\first{}\\
        \bottomrule
    \end{tabular}
    }
\end{table}
\vspace{-0.5em}

\textbf{Zero-shot counting.} The zero-shot modification of DAVE from \Cref{sec:zero-adapt} (denoted here as DAVE$_{\text{0-shot}}$) is compared with the best zero-shot counters LOCA~\cite{djukic_loca}, CounTR~\cite{Liu_2022_BMVC}, RepRPN-C~\cite{ranjan2022exemplar} and RCC~\cite{hobley2022learning}.
The results in Table~\ref{tab:0shotfsc147-results} show that DAVE$_{\text{0-shot}}$ outperforms the state-of-the-art method
LOCA~\cite{djukic_loca}, by a significant margin of 11\% and 7\% MAE on validation and test set, respectively, and outperforms all state-of-the-art in RMSE. 
 
\begin{table}[ht]
    \centering
    \caption{%Comparison with state-of-the-art 
    Zero-shot density-based counting on the FSC147~\cite{famnet}. 
    %\cmnt{[MK] tole gre ven! Note, that methods denoted with * also use text prompts as input.}
    }
    \label{tab:0shotfsc147-results}
    \scalebox{0.9}{
    \begin{tabular}{lllll}
        \toprule
         & \multicolumn{2}{c}{Validation Set} & \multicolumn{2}{c}{Test Set} \\
        \cmidrule(lr){2-3} \cmidrule(lr){4-5}
        Method & MAE  & RMSE & MAE  & RMSE \\
        \midrule
        RepRPN-C~\cite{ranjan2022exemplar} & 29.24 & 98.11 & 26.66 & 129.11 \\
       %  \cmnt{MOVE!!} ZeroClip~\cite{xu2023zero}* &26.93& 88.63 &22.09 &115.17 \\
       % \cmnt{MOVE!!} CLIP-Count~\cite{jiang2023clip}* &18.79&	61.18&17.78	&106.62 \\
        RCC~\cite{hobley2022learning} & 17.49 & 58.81\third{} & 17.12 & 104.53\third{} \\
        CounTR~\cite{Liu_2022_BMVC} & 17.40\second{} & 70.33 & 14.12\first{} & 108.01 \\
        LOCA~\cite{djukic_loca} & 17.43\third{} & 54.96\second{} & 16.22\third{} & 103.96\second{} \\
        DAVE$_{\text{0-shot}}$ &15.54\first{} &52.67\first{}&15.14\second{}&103.49\first{}\\
         % DAVE-clip &15.48&52.65&14.96&103.43\\

        \bottomrule
    \end{tabular}
    }
\end{table}

%\vspace{-0.5em}
\subsection{Detection performance}
\label{sec:detection_perf}

\textbf{Few-shot detection.} Few-shot detectIon performance is evaluated on the FSCD147~\cite{counting-detr} dataset, which
has been recently extended from FSC147~\cite{famnet} by annotating all objects with bounding boxes. We follow the standard evaluation protocol~\cite{counting-detr} with 
Average Precision (AP) and Average Precision at IoU=$50$ (AP50) as the main performance measures. DAVE is compared with the most recent few-shot detection-based counter C-DETR~\cite{counting-detr} as well as adapted few-shot detectors FSDetView~\cite{FSDetView}, AttRPN~\cite{attention-RPN} from~\cite{counting-detr}. 
 
\begin{table}[ht]
\centering
\caption{Detection performance on FSCD147~\cite{counting-detr}.}
\label{tab:fscd147-results}
\scalebox{0.86}{
\begin{tabular}{lllll}
\toprule
  & \multicolumn{2}{c}{Validation Set} & \multicolumn{2}{c}{Test Set} \\
\cmidrule(lr){2-3} \cmidrule(lr){4-5}
Method & AP & AP50 & AP & AP50  \\
\midrule
%FamNet~\cite{famnet}+RR$\dagger$& /&/&9.44&29.73 \\
FSDetView-PB~\cite{FSDetView}&-&-&13.41& 32.99 \\
FSDetView-RR~\cite{FSDetView}&-&-&17.21&33.70 \\
AttRPN-RR~\cite{attention-RPN}&-&-&18.53& 35.87\\
AttRPN-PB~\cite{attention-RPN}&-&-&20.97\third{}&37.19\third{} \\
%\cmnt{SAM}& 20.08&39.02&\textcolor{red}{27.99}&49.17 \\
C-DETR~\cite{counting-detr}& 17.27\second{}&41.90\second{}&22.66\second{}&50.57\second{} \\
%\cmnt{DAVE$_{\text{1-shot}}$} & 18.00 & 52.37 & 19.46 & 55.27 \\
DAVE & 24.20\first{} &61.08\first{} &26.81\first{} &62.82\first{} \\
% DAVE_SAM  & 26.81\first{} &57.10\first{} &36.01\first{} &64.63\first{} \\
% DAVE_SAM_samscore  & 35.18\first{} &64.17\first{} &44.09\first{} &71.96\first{} \\
\bottomrule
\end{tabular}
}
\end{table}

\newpage
\vspace*{-2.7em}
Results in~\Cref{tab:fscd147-results} show that DAVE sets a new state-of-the-art in all measures on both validation and test splits. On the validation split, DAVE outperforms the most recent \mbox{C-DETR}~\cite{counting-detr} by 40\% and 45\% in AP and AP50, respectively, and outperforms C-DETR on the test split by 18\% and 24\% in AP and AP50, respectively.

\begin{figure}[ht]
  \centering
  \includegraphics[width=0.47\textwidth]{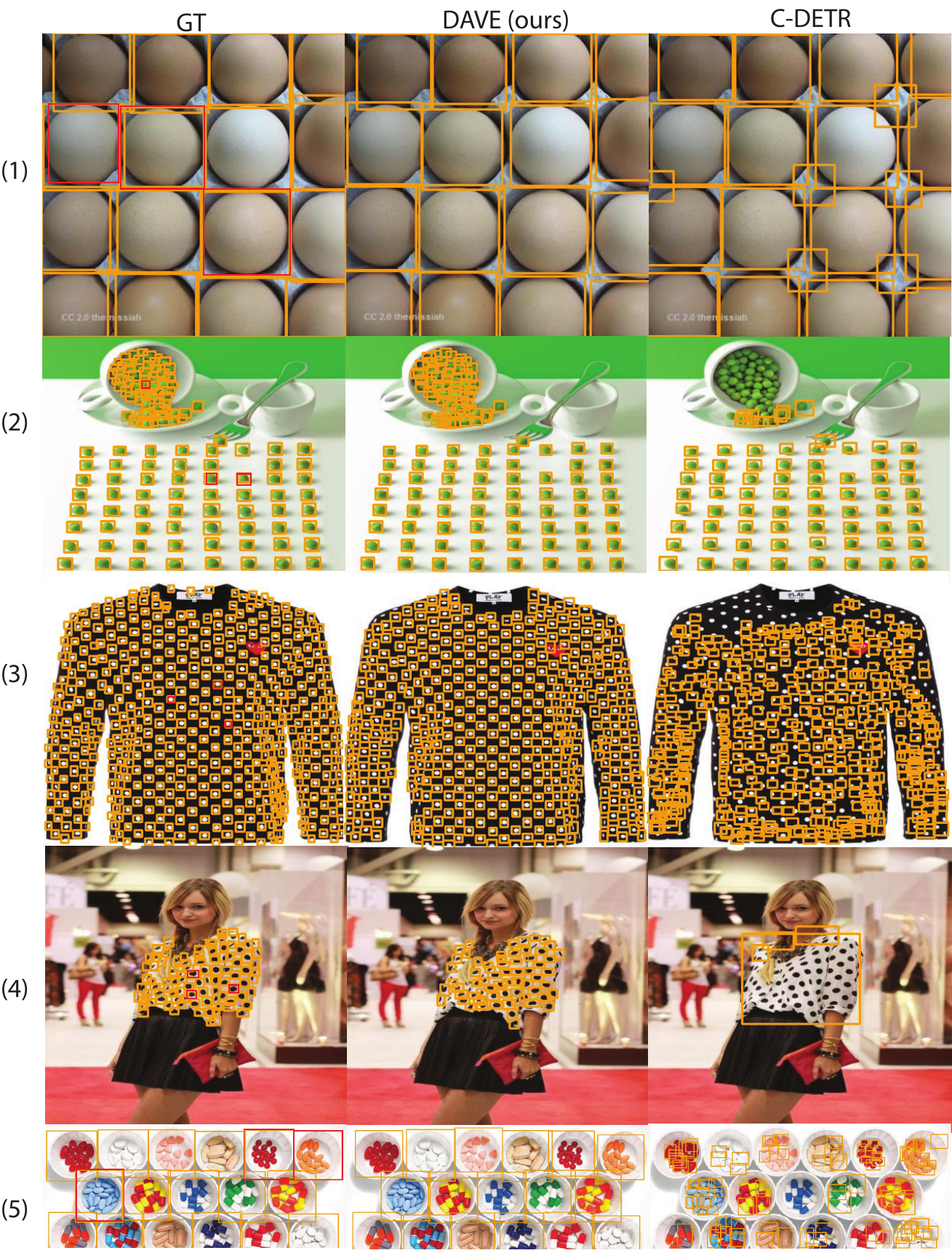} 
  \caption{DAVE localization performance in challenging situations compared with the current best method C-DETR~\cite{counting-detr}.}
  \label{fig:qualitative_det}
\end{figure}

The high AP50 and AP indicate that DAVE retrieves more objects with less false positives, and that localization of the detected objects is more accurate (see~\Cref{fig:qualitative_det}, rows 1 and 2).
DAVE also performs comparatively well in high-density regions with small objects, which are very challenging for the current state-of-the-art (\Cref{fig:qualitative_det}, rows 3 and 4). Compared to the best methods, DAVE better learns the appearance of targets composed of fine-grained objects, leading to improved detections (e.g., bowls of pills in~\Cref{fig:qualitative_det}, row 5).
These results speak of a substantial potential of the detect-and-verify approach for accurate localization. 
% These results speak of a substantial potential of the density-based detect-and-verify approach for accurate localization in detection-based counters. 
  
We further evaluate DAVE on two recent datasets FSCD-LVIS~\cite{counting-detr} and FSCD-LVIS$_\mathrm{uns}$~\cite{counting-detr}. 
Both datasets are created from the LVIS~\cite{gupta2019lvis} dataset containing 
6196 images with 377 classes. In FSCD-LVIS~\cite{counting-detr} dataset, some classes in test set appear also in the training set. The second dataset, FSCD-LVIS$_\mathrm{uns}$~\cite{counting-detr} ensures %(like FSC147) 
that test set does not contain classes observed during training. 
Results in~\Cref{tab:fscd-lvis-results} show that DAVE outperforms the top method C-DETR by 37\% and 55\% w.r.t. AP and AP50, respectively on the FSCD-LVIS. On FSCD-LVIS$_\mathrm{uns}$, DAVE also substantially outperforms the best method by 7\% and 25\% in AP and AP50, respectively. 
 
\begin{table}[ht!]
    \centering
    \caption{Detection on FSCD-LVIS/FSCD-LVIS$_\mathrm{uns}$~\cite{counting-detr} test sets.}
    \label{tab:fscd-lvis-results}
    \scalebox{0.9}{
    \begin{tabular}{llllll}
        \toprule
        & \multicolumn{2}{c}{FSCD-LVIS} & \multicolumn{2}{c}{FSCD-LVIS$_\mathrm{uns}$} \\
        \cmidrule(lr){2-3} \cmidrule(lr){4-5}
         Method & AP  & AP50  & AP  & AP50  \\
        \midrule
        % FamNet~\cite{famnet}+RR$\dagger$ & 0.84 & 2.04 & 0.07 & 0.30\\
        FSDetView-RR~\cite{FSDetView}& 1.96 & 6.70 & 0.89 & 2.38\\
        FSDetView-PB~\cite{FSDetView}& 2.72 & 7.57 & 1.03 & 2.89 \\
        AttRPN-RR~\cite{attention-RPN} & 3.28 & 9.44 & 2.52 & 7.86 \\
        AttRPN-PB~\cite{attention-RPN} & 4.08\third{} & 11.15\third{} & 3.15\third{} & 7.87\third{} \\
        C-DETR~\cite{counting-detr} & 4.92\second{} & 14.49\second{} & 3.85\second{} & 11.28\second{} \\
        DAVE  &6.75\first{}&22.51\first{} & 4.12\first{} & 14.16\first{} \\

        \bottomrule
    \end{tabular}
    }
\end{table}

\textbf{Zero-shot detection.}
To the best of our knowledge, DAVE$_{\text{0-shot}}$ is the first zero-shot method capable of counting and detection. 
We thus compare it with the best counting and detection method C-DETR~\cite{counting-detr}, which however is not zero-shot, since it requires three input exemplars.
Results in \Cref{tab:fscd147-results-0shot} reveal excellent performance of DAVE$_{\text{0-shot}}$. 
On the validation split, it outperforms C-DETR~\cite{counting-detr} by 12\% in AP50 and delivers comparable performance on the test split. While it achieves equally robust detection (AP50) as C-DETR, the localization is slightly less accurate (lower AP).  
Nevertheless, this is a remarkable result, considering C-DETR requires annotated exemplars as input, while DAVE$_{\text{0-shot}}$ does not.
The experiment validates the generality of the proposed detect-and-verify paradigm for all low-shot counting tasks (few- and zero-shot).
%\cmnt{[MK] Tu imamo napačno oznako za metodo? Ker reportamo box-count, metodo pa smo prej poimenovali kot da reporta denisty. Pravim, ker DAVE označujemo DAVEbox, če reporta boxe. Mora bi morali dati box v superscript, kar pomeni, da je ista metoda, le output je drug?}

\begin{table}[ht]
\centering
\caption{
%DAVE$_{\text{0-shot}}$ detection performance compared with a few-shot detection counter on FSCD147~\cite{counting-detr}.
Without requiring exemplars, DAVE$_{\text{0-shot}}$ performs on par or better than 
C-DETR with three input exemplars.
%{\color{orange} DAVE$_{\text{0-shot}}$ achieves comparable (or better) performance on FSCD147~\cite{counting-detr} without any given exemplar, compared to C-DETR with three input exemplars.} 
}
\label{tab:fscd147-results-0shot}
\scalebox{0.86}{
\begin{tabular}{lllll}
\toprule
  & \multicolumn{2}{c}{Validation Set} & \multicolumn{2}{c}{Test Set} \\
\cmidrule(lr){2-3} \cmidrule(lr){4-5}
Method & AP & AP50  & AP  & AP50  \\
\midrule
%FamNet~\cite{famnet}+RR$\dagger$& /&/&9.44&29.73 \\
% FSDetView~\cite{FSDetView}+PB$\dagger$&/&/&13.41& 32.99 \\
% FSDetView~\cite{FSDetView}+RR$\dagger$&/&/&17.21&33.70 \\
% Att-RPN~\cite{attention-RPN}+RR$\dagger$&/&/&18.53& 35.87\\
% Att-RPN~\cite{attention-RPN}+PB$\dagger$&/&/&20.97\third{}&37.19\third{} \\
%\cmnt{SAM}& 20.08&39.02&\textcolor{red}{27.99}&49.17 \\
C-DETR (\textit{3-shot})~\cite{counting-detr}& 17.27\first{}&41.90\second{}&22.66\first{}&50.57\first{} \\
DAVE$_{\text{0-shot}}$ & 16.31\second{} &46.87\first{} &18.55\second{} &50.08\second{} \\
%\cmnt{DAVE$_{\text{prm}}$} & 16.32 &46.91 &18.64 & 50.24\\
\bottomrule
\end{tabular}
}
\end{table}

\textbf{Few-shot detection counting.}
The previous experiments analyzed the accuracy of detections. To further analyze the detection capability, we measure the accuracy of count estimation when approximated by the number of detected bounding boxes. In the following, we use the superscript  DAVE$^{\text{box}}$ to distinguish the results from the density-based count estimation.
Results are reported in~\Cref{tab:fsc147-box_count-results}. DAVE$^{\text{box}}$ outperforms all state-of-the-art by a significant margin, in particular, it outperforms C-DETR by 38\% in MAE and 40\% in RMSE. This further confirms the remarkable detection performance compared to the most recent detection-based methods. 
Note that the DAVE$^{\text{box}}$ not only outperforms all detection-based counters, but also all published density-based counters in terms of MAE, including LOCA~\cite{djukic_loca} (\Cref{tab:fsc147-results}), which have been up to now unchallenged by the detection-based counters.

\begin{table}[ht]
\centering
\caption{Few-shot detection-based counting on FSC147~\cite{famnet}.}
\label{tab:fsc147-box_count-results}

\resizebox{1 \linewidth}{!}{
\begin{tabular}{lllll}
\toprule
  & \multicolumn{2}{c}{Validation Set} & \multicolumn{2}{c}{Test Set} \\
\cmidrule(lr){2-3} \cmidrule(lr){4-5}
Method & MAE & RMSE & MAE & RMSE  \\
\midrule
FSDetView-RR~\cite{FSDetView}&-&-& 37.83 &146.56  \\
FSDetView-PB~\cite{FSDetView}&-&-& 37.54 & 147.07  \\
AttRPN-RR~\cite{attention-RPN}&-&-& 32.70 & 141.07\third{}  \\
AttRPN-PB~\cite{attention-RPN}&-&-& 32.42\third{}& 141.55  \\
C-DETR~\cite{counting-detr}& 20.38\second{}&82.45\second{} &16.79\second{}& 123.56\second{} \\
DAVE$^{\text{box}}$ &9.75\first{}& 40.30\first{}& 10.45\first{}  & 74.51\first{}  \\
\bottomrule
\end{tabular}}
\end{table}

 \vspace{-1em}
\subsection{Ablation study} \label{sec:ablation}

\textbf{{Impact of mixed-class training.}} 
We first verify whether false positives in state-of-the-art methods could be reduced by simply training on images with multiple object categories.
The current top low-shot counter LOCA~\cite{djukic_loca} is thus retrained on multi-class images, in which a FSCD147~\cite{counting-detr} training image is concatenated with another image containing objects from a different class for hard negative training examples, as described in \Cref{sec:training}. 
We denote this version by LOCA$_{\text{mul}}$ and also include CounTR~\cite{Liu_2022_BMVC} in the comparison since it already applies such a training setup. 
% Results in~\Cref{tab:FSC-stitched} are reported on the dataset FSCD147 and on a subset FSCD147$_\mathrm{mul}$, which is composed of
% images that contain many objects different from the exemplar class, thus emphasizing robustness to other-class objects.  
We also construct a subset of FSCD147 composed of images containing objects from different classes\footnote{Images were obtained from the test and evaluation splits of FSCD147.}
(denoted as FSCD147$_\mathrm{mul}$), to expose the sensitivity of a counting method to other-class objects. 
Table~\ref{tab:FSC-stitched} shows that the counting performance of LOCA on FSCD147$_\mathrm{mul}$ is significantly lower compared to FSCD147, confirming that multi-class images are highly challenging. 
Training LOCA on multi-class images (LOCA$_\mathrm{mul}$) 
substantially reduces the average error on FSCD147$_\mathrm{mul}$ by 43\%, but increases it by 28\% on FSC147. 
This is likely due to LOCA$_{\text{mul}}$ compensating for the improved multi-class performance by a reduced overall performance. 
However, DAVE demonstrates excellent performance on both datasets and also consistently outperforms both LOCA versions and CounTR by large margins. 
 
\textbf{{Impact of cluster selection.}}
We compare the prompt-based DAVE$_\text{prm}$ with DAVE$_\text{0-shot}$ to demonstrate the impact of the cluster selection method. % in the zero-shot setup. 
Notably, DAVE$_\text{prm}$ selects the clusters by comparing them with text prompts, while DAVE$_\text{0-shot}$ applies majority voting (Section~\ref{sec:zero-adapt}). 
While the performance of the two methods is comparable on average, DAVE$_\text{prm}$ substantially outperforms on FSCD147$_\text{mul}$. 
This result is presented in Table~\ref{tab:FSC-stitched} (bottom) and indicates that prompt-based cluster selection is particularly important on images with multiple classes to resolve the object category ambiguity.
\vspace{-0.2em}
\begin{table}[ht]
    \centering
    \caption{
    % Performance in presence of other-class objects.
    %Analysis of multiple classes presence (FSCD147$_\mathrm{mul}$)
    Performance in presence of objects from multiple classes. 
    %(FSCD147$_\mathrm{mul}$).
    %Comparison on FSCD147 and FSCD147$_\mathrm{mul}$.
    }
    \label{tab:FSC-stitched}
    \resizebox{\linewidth}{!}{
    \begin{tabular}{llllll}
        \toprule
                  & \multicolumn{3}{c}{FSCD147} & \multicolumn{2}{c}{FSCD147$_{\text{mul}}$} \\
        \cmidrule(lr){2-4} \cmidrule(lr){5-6}
         & MAE($\downarrow$) & RMSE($\downarrow$) & AP50($\uparrow$) & MAE($\downarrow$) & RMSE($\downarrow$)  \\
        \midrule
        \rowcolor{Gray}
        LOCA~\cite{djukic_loca} &               10.79\second{}  &56.97\second{} &        -          & 21.28             & 43.67             \\
        \rowcolor{Gray}
        LOCA$_{\text{mul}}$~\cite{djukic_loca} & 12.63           & 78.95         &        -          & 13.25\second{}    & 22.57\second{}    \\
        \rowcolor{Gray} 
        CounTR~\cite{Liu_2022_BMVC} &           11.95\third{}   &91.23\third{}  &        -          & 14.56\third{}     & 27.41\third{}    \\
        \rowcolor{Gray} 
        % C-DETR~\cite{counting-detr}                                                               & 23.09             &  30.09 \\
        DAVE&                                   8.66\first{}    &32.36\first{}  & 61.08\first{}     &3.05\first{}       &4.94\first{}       \\
        \arrayrulecolor{gray}
        \midrule
        
        % DAVE$_{\overline{\text{R}}}$          & \gc{8.97}       &\gc{28.12}     & 53.57             & 3.27              & 4.98             & \\
        % DAVE$_{\text{cat}}$                   & \gc{8.99}       &\gc{28.18}     & 51.72             & 2.85              & 4.38             & \\
        % DAVE$_{\text{sum}}$                   & \gc{8.95}       &\gc{28.13}     & 55.54             & 2.71              & 4.29             & \\
        % DAVE$_{\overline{\text{prj}}}$        &     9.41        &    29.91      & 60.85             & 20.91             & 43.67            & \\
        DAVE$_{\text{0-shot}}$ &                 15.54           &103.49         & 50.08             &12.86              &23.21             \\

        DAVE$_{\text{prm}}$ &                 14.90           &103.42         & 50.24             &6.46               &10.72             \\

       % \rowcolor{Gray}
       % DAVE$_{\text{3-shot}}$&                 8.66             &32.36          & 61.08            &3.05                &4.94             & \\

       \arrayrulecolor{black}
        \bottomrule
    \end{tabular}
    }
\end{table}
\vspace*{-1.5em}
\begin{table}[ht]
    \centering
    \caption{
    % Ablation analysis on FSCD147~\cite{famnet}.
    DAVE architecture analysis on FSCD147~\cite{famnet}.
    }
    \label{tab:ablation}
    \resizebox{0.9\linewidth}{!}{
    \begin{tabular}{lllll}
        \toprule
         & MAE($\downarrow$) & RMSE($\downarrow$) & AP($\uparrow$)& AP50($\uparrow$)  \\
        \midrule
        DAVE                                  & \textbf{8.91}&  \textbf{28.08 }  &  \textbf{24.19 }&  \textbf{61.08} \\
        DAVE$_{\overline{\phi}}$        &     9.41        &    29.91   & 24.11  & 60.85             \\
        DAVE$_{\overline{\text{R}}}$          & 8.97       &28.12 & 19.50   & 53.57              \\
        DAVE$_{\text{cat}}$                   & 8.99      &28.18  & 18.49  & 51.72           \\
        DAVE$_{\text{sum}}$                   & 8.95       &28.13 & 20.74  & 55.54             \\
       \arrayrulecolor{black}
        \bottomrule
    \end{tabular}
    }
\end{table}
 
\textbf{{Architecture design.}}
Finally, we evaluate the DAVE architectural design decisions. 
First, we analyze the impact of using the prototype correlation response tensor $\tilde R$ in the box regression step. 
\Cref{tab:ablation} shows that removing $\tilde R$ (DAVE$_{\overline{\text{R}}}$) results in a substantial drop of 12\% and 9\% in AP and AP50. This verifies the importance of fusion with $\tilde R$, which contains size and shape information of the selected objects, considerably improving the localization accuracy of DAVE detections. 
In particular, for target objects composed of smaller objects, this information is crucial for accurate bounding box prediction (\Cref{fig:qualitative_det}, last row). 
To evaluate the importance of the feature fusion module (FFM), we replace it with sum (DAVE$_{{\text{sum}}}$), and concatenation (DAVE$_{{\text{cat}}}$). Both replacements result in a detection performance drop of 9\% and 15\% AP, respectively. 
To validate the importance of robust appearance features in the verification stage, we remove the feature projection network $\phi(\cdot)$ and perform clustering directly on the backbone features (DAVE$_{\overline{\phi}}$).
The errors of DAVE$_{\overline{\phi}}$ increase by 6\% in MAE and 7\% RMSE.

\ifdefined\ArxivCompilePrepub
\else

\textbf{Limitations.} DAVE outputs detections (i.e., bounding boxes), as well as total counts estimated from the density. To expose limitations, we inspect the discrepancy between the total count estimates and the number of detections with respect to the number of objects in the image (\Cref{fig:limitations}). The discrepancy is most apparent for images with very large object counts, which typically contain many small objects packed together (i.e., extremely dense regions). Further error reductions are thus expected by improving DAVE detection stage in the presence of extreme density. The limitation is common to all low-shot counters, and we defer this to future research. 

\begin{figure}[ht]
  \centering
  \includegraphics[width=0.45\textwidth]{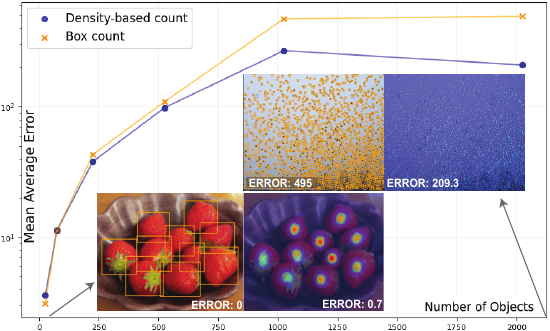} 
  \caption{DAVE density-based and box-count accuracy with respect to the number of objects in the image.
  }
  \label{fig:limitations}
\end{figure}

\fi 

%%%%%%%%%%%%%%%%%%%%%%%%%%%%%%%%%%%%%%%%%%%%%%%%%%%%%%%%%%%%%%%%%%%%%%%%%%%%%%%%
\vspace{-2em}
\section{Conclusion} \label{ch:conclusions}
\vspace{-0.4em}
%%%%%%%%%%%%%%%%%%%%%%%%%%%%%%%%%%%%%%%%%%%%%%%%%%%%%%%%%%%%%%%%%%%%%%%%%%%%%%%%

We presented a novel low-shot object counting and detection method DAVE, that narrows the performance gap between density-based and detection-based counters. DAVE spans the entire low-shot spectrum, also covering text-prompt setups, and is the first method capable of zero-shot detection-based counting.
This is achieved by the novel detect-and-verify paradigm, which increases the recall as well as precision of the detections. 
 
Extensive analysis demonstrates that DAVE sets a new state-of-the-art in total count estimation, as well as in detection accuracy on several benchmarks with comparable complexity to related methods, running 110ms/image. 
In particular, DAVE outperforms the long-standing top low-shot counter~\cite{djukic_loca}, as well as the recent detection-based counter~\cite{counting-detr}.
%and outperforms the most recent detection-based counter~\cite{counting-detr} by $\sim$15\% \cmnt{AP50}.
In a zero-shot setup, DAVE outperforms all density-based counters and delivers detections on par with the most recent few-shot counter that requires at least few annotations. 
DAVE also sets a new state-of-the-art in prompt-based counting.
In our future work, we plan to explore interactive counting with the human in the loop and improve detection in extremely dense regions.

{\small
\noindent \textbf{Acknowledgements.} This work was supported by Slovenian research agency program P2-0214 and projects J2-2506, L2-3169 and Z2-4459.
}

{
    \small
    \bibliographystyle{ieeenat_fullname}
    \bibliography{main}
}


\end{document}